\documentclass{publ}
\usepackage[T1]{fontenc}
\usepackage{amsmath}
\usepackage{amssymb}
\usepackage{cite}  %
\usepackage{pifont}
\usepackage{microtype}
\usepackage{float}
\usepackage{lipsum} 
\usepackage{subfig}
\usepackage{algorithm}
\usepackage{algpseudocode}
\usepackage{multirow}
\usepackage{siunitx}
\BibtexOrBiblatex
\electronicVersion
\PrintedOrElectronic
\usepackage[pdftex]{graphicx} 
\usepackage[a4paper, margin=0.65in, top=0.9in, bottom=1.23in]{geometry}
\usepackage{lineno}
\usepackage{fancyhdr}
\usepackage{cleveref}
\usepackage{xspace}
\newcommand{\name}{\textsc{ProteusNeRF}\xspace}
\newcommand{\Context}{3D-Aware Image Context\xspace}
\newcommand{\context}{3D-aware image context\xspace}
\newcommand{\triplanelite}{TriPlaneLite\xspace}
\newcommand{\mask}{\ensuremath{\mathcal{M}_\text{sel}}\xspace}
\newcommand{\mysec}[1]{Section~\ref{#1}\xspace}
\newcommand{\appTime}{10 seconds\xspace}
\newcommand{\geoTime}{70 seconds\xspace}

\title[Nerf Editing using 3D-aware Image Context]%
      {\name: Fast Lightweight NeRF Editing \\using 
\Context}

\author{
    \textbf{Binglun Wang} \hspace{0.5cm}
    \textbf{Niladri Shekhar Dutt} \hspace{0.5cm}
    \textbf{Niloy J. Mitra}\\ 
    \large{\{binglun.wang.22, niladri.dutt.22\}@ucl.ac.uk} \\\\
    \large{University College London}
    \\\\\large{\href{https://proteusnerf.github.io}{https://proteusnerf.github.io}}
}

\begin{document}

 \teaser{
 \includegraphics[width=\linewidth]{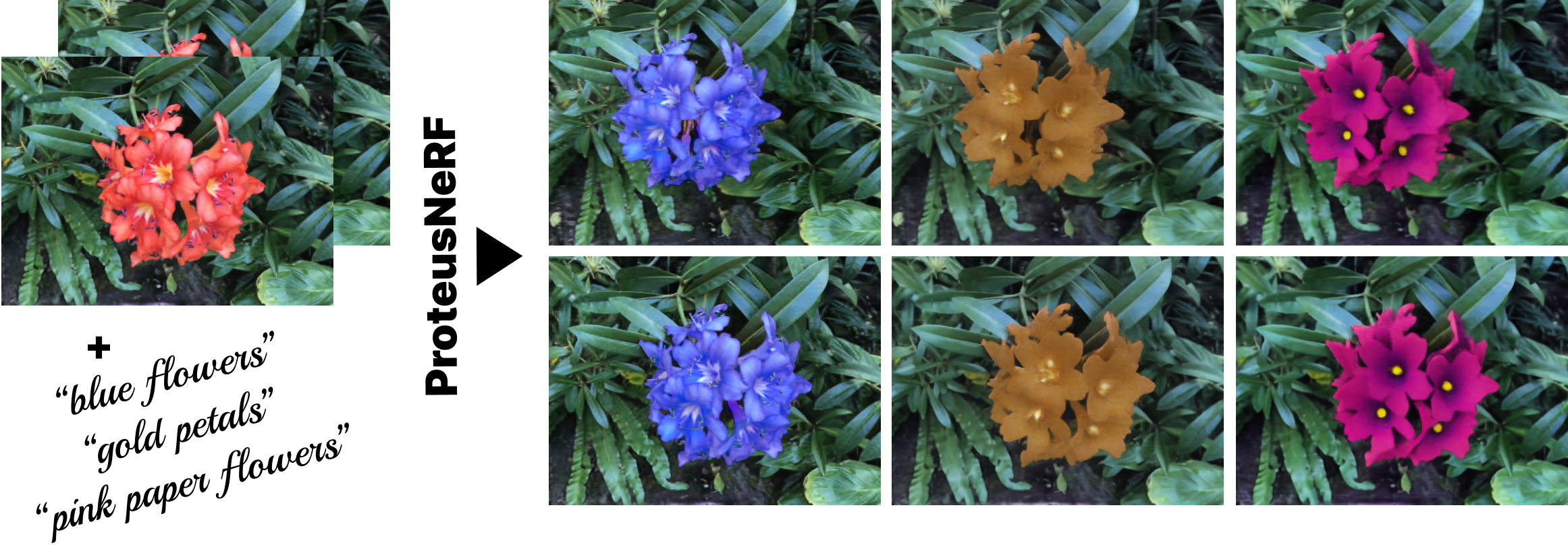}
  \centering
   \caption{
We present \name, a fast and lightweight framework for editing NeRF assets via existing image manipulation tools, traditional or generative. We enable this by a novel \context that allows linking edits across multiple views. Here, we show Nerf edits (original and novel views) using text-guided edits. These edits take 10-70 seconds.
}
 \label{fig:teaser}
\vspace*{.1in}
}

\maketitle

\begin{abstract}  

Neural Radiance Fields~(NeRFs) have recently emerged as a popular option for photo-realistic object capture due to their ability to faithfully capture high-fidelity volumetric content even from handheld video input. Although much research has been devoted to efficient optimization leading to real-time training and rendering, options for interactive editing NeRFs remain limited. 
We present a very simple but effective neural network architecture that is fast and efficient while maintaining a low memory footprint. This architecture can be incrementally guided through user-friendly image-based edits. 
Our representation allows straightforward object selection via semantic feature distillation at the training stage. More importantly, we propose a local \context to facilitate view-consistent image editing that can then be distilled into fine-tuned NeRFs, via geometric and appearance adjustments. 
We evaluate our setup on a variety of examples to demonstrate appearance and geometric edits and report 10-30$\times$ speedup over concurrent work focusing on text-guided NeRF editing. Video results and code can be found on our project webpage at \href{https://proteusnerf.github.io}{https://proteusnerf.github.io}.


\end{abstract}  

\section{Introduction}

Neural Radiance Fields~(NeRFs) \cite{mildenhall2020nerf} have rapidly emerged as one of the most popular volumetric representations for casual capture of 3D objects. Benefiting from a significant body of improvements to the original formulation, it is now possible to optimize NeRFs in a matter of minutes and generate photo-realistic novel views at interactive framerates.

In this work, we focus on interactively editing such NeRF assets while preserving their original volumetric representation (note that we do not focus on approaches where one first distills NeRFs into textured surface meshes). A good editing system should be simple to use, expressive, light-weight, and encourage interactive manipulation. One approach is to enclose NeRF assets into volumetric cages \cite{xu2022deforming} and then edit the underlying volumes using cage-based deformation setups. NeRFshop~\cite{jambon2023nerfshop} presents an impressive realization of this workflow where users select and prescribe (deformation) handles in 3D. However, unlike cage based methods which focus on editing the geometry of the NeRFs, we focus on image-based workflows, primarily targeting appearance changes. 

Inspired by the recent success of large text or image-conditioned generative models~\cite{stablediffusion,ramesh2022hierarchical} that directly produce NeRF assets \cite{metzer2023latent, poole2022dreamfusion, jain2022zero}, we investigate the feasibility of an entirely image-based NeRF editing framework. In our setup, users only interact with assets using image-based interfaces for selection and editing. This allows users to benefit from  existing image editing tools, both traditional and recent generative methods, without having to learn a new editing setup.  

However, to realize the above goal, we first need to solve a few problems: (i)~enable object selection, (ii)~perform synchronized multi-view image edits, and (iii)~update the NeRF assets. We demonstrate that achieving all the outlined problems is possible with feature distillation, a very simple adaptation to existing architectures, and a novel 3D-aware image context, respectively. Technically, once objects are selected, our method alternates between two phases: (a)~using the pre-trained NeRF to produce 3D-aware image context that can be edited using existing image manipulation frameworks -- traditional or generative, and (b)~interleaving between geometric and appearance updates to distill the image edits into an edited NeRF asset. 
The resultant edits are lightweight~(4-36KB/edit or 32MB/edit) and fast (\appTime-\geoTime/edit) to realize. 
We encode the edits as residual updates, which being localized and lightweight can be stored as edit tokens. Such edits  can be enabled or disabled in a post-edit setup, akin to layered edit updates in traditional image editing workflows. 

In summary, we introduce a simple, fast, and lightweight NeRF editing setup. We demonstrate the effectiveness of the proposed framework on a diverse set of edit scenarios. In our experiments, we achieve 10-30$\times$ speedup over concurrent generative NeRF edit setups \cite{haque2023instruct}.

\section{Related Work}

\noindent\textbf{Novel view synthesis.} Neural Radiance Fields (NeRFs)~\cite{mildenhall2020nerf} have been highly successful at generating photo-realistic novel views of a 3D scene by optimizing an object-specific MLP to simultaneously model geometry and appearance. However, training a global MLP as a NeRF representation is a slow process and therefore several improvements have been proposed~\cite{garbin2021fastnerf,muller2022instant,hedman2021baking,yu2021plenoctrees}. For example, 
ReluFields~\cite{ReluField_sigg_22} use a uniform grid, while 
Plenoxels~\cite{yu_and_fridovichkeil2021plenoxels} and DVGO~\cite{SunSC22} utilize a sparse voxel grid for 3D scene reconstruction using distributed and localized representations. More relevant to ours, structured representation in the form of collection of 2D representations have been proposed to enable efficient storage. For example, triplane representations such as K-planes~\cite{kplanes_2023}, EG3D~\cite{eg3d}, and TensoRF~\cite{Chen2022ECCV}, having fewer variables to optimize (i.e., storage requirement being quadratic versus cubic in grid resolution), and can achieve high quality results within minutes.

\begin{figure}[t!]
    \centering
    \includegraphics[width=\columnwidth]{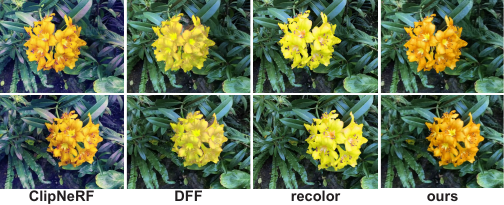}
    \caption{
    Visual comparison of color editing. CLIP-NeRF~\cite{wang2022clip} sees color bleeding into the global scene, and DFF~\cite{kobayashi2022decomposing} shows undesirable color changes in the pistil and unnatural color gradient. Our approach matches the impressive results of RecolorNeRF~\cite{gong2023recolornerf}, while offering a more intuitive and flexible framework and taking a fraction of time for editing (10 seconds vs 2-3 minutes for RecolorNeRF).}
    \label{fig:motivation}
\end{figure}

\noindent\textbf{Image editing.} In traditional image manipulation, decades of mature research exists behind image editing tools (e.g., Gimp, Photoshop) that can perform operations such as tone mapping, color, contrast, hue changes, etc. In contrast, recent breakthroughs in generative models such as VAEs \cite{kingma2013auto, harshvardhan2020comprehensive}, GANs \cite{goodfellow2020generative, wang2021generative}, and diffusion models \cite{ho2020denoising, croitoru2023diffusion} have led to remarkable results in image editing by utilizing image priors distilled from large image collections. For instance, DragGAN \cite{pan2023drag} enables image deformations by allowing a user to move a set of handle points towards target points, while InsturctPix2Pix \cite{brooks2023instructpix2pix} and ControlNet \cite{controlnet} allow image editing using simple text prompts.

\begin{table}[b!]
    \centering
    \caption{
    \textbf{Comparison wrt related/concurrent works.} 
    Limited options are available to explore geometric and appearance edits of NeRFs interactively~(Fast). Ours strikes a balance between being expressive and lightweight, while being able to make limited geometric changes~(Geo.), add small geometric details~(Add.), delete selected objects~(Del.), or make larger appearance changes~(App.).}
    \small
    \begin{tabular}{p{3.6cm}|p{0.45cm}|p{0.45cm}|p{0.45cm}|p{0.45cm}|p{0.4cm}}
         Method & App. & Geo. & Del. & Add & Fast\\ 
         \hline \hline
        PaletteNeRF\cite{kuang2023palettenerf}& 
        \hspace{0.4em}\ding{52} & \hspace{0.4em}\ding{56}  & \hspace{0.4em}\ding{56} & \hspace{0.4em}\ding{56} & \hspace{0.4em}\ding{56} \\
        RecolorNeRF\cite{gong2023recolornerf}& 
        \hspace{0.4em}\ding{52} & \hspace{0.4em}\ding{56}  & \hspace{0.4em}\ding{56} & \hspace{0.4em}\ding{56} & \hspace{0.4em}\ding{52} \\
        ICE-NeRF\cite{lee2023ice}& 
        \hspace{0.4em}\ding{52} & \hspace{0.4em}\ding{56}  & \hspace{0.4em}\ding{56} & \hspace{0.4em}\ding{56} & \hspace{0.4em}\ding{52} \\
        DreamEditor \cite{zhuang2023dreameditor} & 
        \hspace{0.4em}\ding{52} & \hspace{0.4em}\ding{52}  & \hspace{0.4em}\ding{56} & \hspace{0.4em}\ding{52} & \hspace{0.4em}\ding{56} \\ 
        CLIPNeRF  \cite{wang2022clip}& 
        \hspace{0.4em}\ding{52} & \hspace{0.4em}\ding{52}  & \hspace{0.4em}\ding{56} & \hspace{0.4em}\ding{56} & \hspace{0.4em}\ding{56} \\
        NeRFshop \cite{jambon2023nerfshop}& 
        \hspace{0.4em}\ding{56} & \hspace{0.4em}\ding{52}  & \hspace{0.4em}\ding{52} & \hspace{0.4em}\ding{52} & \hspace{0.4em}\ding{52} \\
        InstructN2N.~\cite{haque2023instruct} & 
        \hspace{0.4em}\ding{52} & \hspace{0.4em}\ding{52}  & \hspace{0.4em}\ding{56} & \hspace{0.4em}\ding{52} & \hspace{0.4em}\ding{56} \\ 
        DFF\cite{kobayashi2022decomposing}& 
        \hspace{0.4em}\ding{52} & \hspace{0.4em}\ding{52}  & \hspace{0.4em}\ding{52} & \hspace{0.4em}\ding{52} & \hspace{0.4em}\ding{56} \\
        InpaintNeRF360 \cite{wang2023inpaintnerf360} & 
        \hspace{0.4em}\ding{52} & \hspace{0.4em}\ding{56}  & \hspace{0.4em}\ding{52} & \hspace{0.4em}\ding{56} & \hspace{0.4em}\ding{56} \\ 
        \name (Ours) & 
        \hspace{0.4em}\ding{52} & \hspace{0.4em}\ding{52}  & \hspace{0.4em}\ding{52} & \hspace{0.4em}\ding{52} & \hspace{0.4em}\ding{52} \\
    \end{tabular}
    \label{tab:compareRelatedWk}
\end{table}

\noindent\textbf{Editing NeRFs.} Recent advances have primarily focused on text-to-3D conditional generation \cite{metzer2023latent, poole2022dreamfusion, jain2022zero} or artistic stylization of existing NeRF assets \cite{nguyen2022snerf, chiang2022stylizing, huang2022stylizednerf, huang2021learningstyle}. Methods that focus on directly generating 3D content from text lack fine-grained control of the generated scene and hence synthesize arbitrary scenes, which does not help to make changes to an existing or captured scene. Stylization methods such as CLIP-NeRF \cite{wang2022clip}, NeRF-Art \cite{wang2022nerf}, and Blending NeRF \cite{song2023blending} edit a scene by optimizing global stylistic similarity of reconstructed views and an edit text prompt in the CLIP latent space \cite{radford2021learning}. These methods focus on changing global scenes; therefore, local selection methods such as NVOS~\cite{nvos} can be used for 3D segmentation. To enable local edits, Distilled Feature Fields \cite{kobayashi2022decomposing} and Neural Feature Fusion Fields \cite{tschernezki2022neural} propose to distill features from large scale 2D pre-trained models into 3D to perform selective editing of regions.
NeRF Analogies \cite{fischer2024nerf} utilizes distilled features to transfer the appearance of a source object to a target object.
DreamEditor \cite{zhuang2023dreameditor} first distills the NeRF into a mesh-based field and then uses score distillation \cite{poole2022dreamfusion} to optimize any local edit.
PaletteNeRF \cite{kuang2023palettenerf}, RecolorNeRF\cite{gong2023recolornerf}, and ICE-NeRF \cite{lee2023ice} can produce color changes by decomposing a scene into multiple color palettes but the editing landscape is very limited. Moreover, the color based selection is not very robust compared to feature based selection and can hence produce unwanted recoloring. 

More closely related to ours, in order to enable a more intuitive interface for editing, are the concurrent works of InpaintNeRF360 \cite{wang2023inpaintnerf360} and InstructNeRF2NeRF \cite{haque2023instruct}. They propose to use natural language as instruction to guide the editing process. InpaintNeRF360 \cite{wang2023inpaintnerf360} updates the NeRF scene with inpainted images using segmentation masks obtained from point-based prompts and accurate bounding boxes using depth information from the pre-trained NeRF. InstructNeRF2NeRF \cite{haque2023instruct} iteratively updates the dataset by modifying the rendered images from the pre-trained NeRF using InstructPix2Pix \cite{brooks2023instructpix2pix}. To facilitate geometric editing, Deforming-NeRF \cite{xu2022deforming} first encloses the foreground object in a cage and then deforms it by maneuvering the vertices of the cage. NeRFshop \cite{jambon2023nerfshop} improves this paradigm to enable interactive object selection using scribbles.

\noindent Although existing methods produce good edited results, to maintain multi-view consistency, re-training the NeRF remains a computationally expensive problem, taking 30 minutes to 2 hours per edit. Taking inspiration from TextMesh~\cite{tsalicoglou2023textmesh}, which combines renders from four orthogonal views to process them using diffusion jointly, we create a 3D-aware image-grid context to enable fast re-training via a novel \triplanelite architecture while maintaining view-consistency. As shown in \Cref{tab:compareRelatedWk}, our method can handle a variety of edits while only taking a fraction of the time compared to existing methods. Note that many of these are concurrent/unpublished works and we could not get code access for comparison. 

\section{Background}
\subsection{Neural Radiance Fields Revisited}

Neural Radiance Fields (NeRFs) \cite{mildenhall2020nerf} generate photo-realistic novel views of a 3D scene by representing it as a radiance field approximated by a neural network, usually an MLP. It takes as input a 3D spatial location ($\mathbf{p} := (x,y,z)$) and a view direction ($\theta, \phi$), and produces as output the corresponding color, $\mathbf{c}_{3D}$ (i.e., $R, G, B$) and volume density ($\sigma$). To render a pixel in an image, NeRF projects rays from the camera, and then samples points along the rays in 3D space. Using classical volumetric rendering \cite{porter1984compositing}, the colors and densities are converted into pixel colors by integrating the sampled color  of 3D points $i$ along the ray as, 
\begin{equation} \label{nerf-eqn}
c_{2D} := \sum_i T_i\alpha_i {\mathbf{c}}_{3D}^i,
\end{equation} 

Where, $\alpha$ is opacity, $\alpha_i = 1-\exp(-\sigma_i\delta_i)$; $T$ is the transmittance, $T_i = \prod_j^{i-1}(1-\alpha_j)$ is the probability of the ray passing through the volume up to the sampled 3D point $i$. In other words, for each sampled 3D point $i$ along the ray, $T_i$ simulates the probability that the ray can pass through sampled points $j$ before the point $i$ in the ray direction; $\alpha_i$ simulates the probability that the ray 'hits' or cannot pass through the point $i$ \cite{2022eccvtutorialEncoding}; ${\mathbf{c}}_{3D}^i$ denotes the color on the $i$-th point on the ray and $\delta_i$ is the distance between sampled points along the ray. 

\begin{figure*}[t!]
  \centering
  \includegraphics[width=\textwidth]{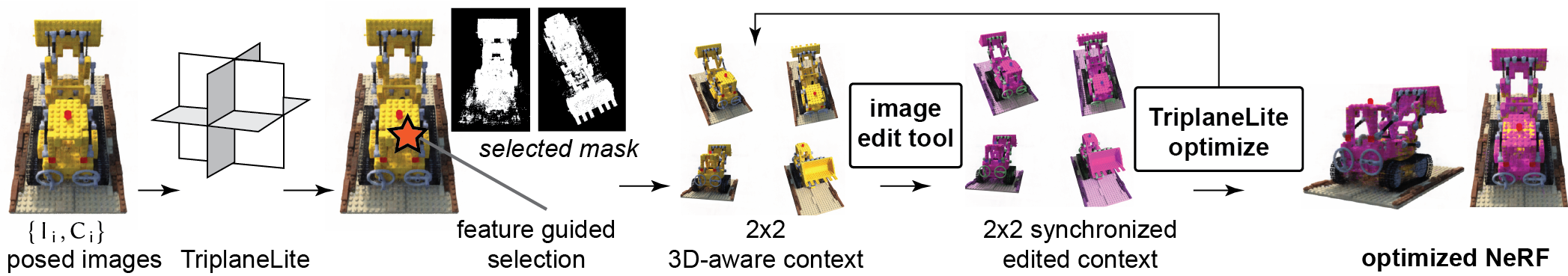}
  \caption{We present \name that takes in a set of posed images and encodes it as feature-distilled NeRF in a TriplaneLite representation. The user can easily select a part (yellow legos) that gets converted to a 3D mask \mask. We generate a novel \context that allows editing via imaging tools while still producing view-coherent edits. This edited context is then converted back to view-consistent NeRFs by fine-tuning the TriplaneLite. The context image is updated and the process is iterated (2-3 times in our examples). Editing, primarily appearance editing, runs at interactive framerates. }
  \label{fig:overview}
\end{figure*}

\subsection{Radiance Feature Fields}
Radiance Feature Fields~\cite{kobayashi2022decomposing, tschernezki22neural} extends NeRFs to capture semantic features to enable the selection of specific regions in a scene by distilling features from large scale self-supervised image models such as DINO \cite{dino} into 3D feature fields ($f_{sem}$). To render the color ($c_{2D}$) and semantic features ($f_{2D}$) of a pixel, it follows volume rendering as, 
\begin{equation} \label{volume-features}
\begin{split}
c_{2D} &:= \sum_i T_i\alpha_i {\mathbf{c}}_{3D}^i \\
f_{2D} &:= \sum_i T_i\alpha_i {\mathbf{f}}_{sem}^i,
\end{split}
\end{equation}
where we additionally optimize/distill volumetric features ${\mathbf{f}}_{3D}$ at each location in space.

\subsection{Representing NeRFs as Tri-planes}
Storing a distributed NeRF in a volume runs into cubic storage space in the resolution of the underlying grid. We can significantly reduce the training time and memory footprint for NeRFs by representing it as a tri-plane \cite{eg3d}. This approach maps a 3D spatial location ($\mathbf{p}:=(x,y,z)$) to 3 hidden feature vectors as inputs to the NeRF by interpolating among three distinct feature planes, $P_{xy}$, $P_{xz}$, and $P_{yz}$, via orthographically projecting $\mathbf{p}$ to the three planes, and adding (or concatenating) the feature vectors. K-planes~\cite{kplanes_2023} alternately restructures the information by multiplying the feature vectors instead of adding them.

\section{\name}

\paragraph*{Overview.}
Starting from a set of posed images (i.e., images with associated cameras $\{I_i, C_i\}_{i=1:n}$), we first optimize a NeRF represented using our \triplanelite representation (\mysec{subsec:triplanelite}). 
The user then chooses a camera view and indicates an image patch from the rendered frame to select any object she wishes to modify. 
Based on the image selection, we use the distilled semantic features to automatically select an object region, denoted by a 3D mask \mask, relying on similarity in the tri-plane feature field. 
We then render images from sampled views in the training dataset to form a 2$\times$2 \textit{\context} to provide  guidance (\mysec{subsec:gridContext}). The user then uses image editing tools, either a traditional image editor (e.g., Gimp, Photoshop) or a text-guided generative editor (e.g., InstructPix2Pix~\cite{brooks2023instructpix2pix}), to edit the guidance image. 
For edits involving only appearance changes, thanks to our \triplanelite architecture, we train a residual MLP with a minimal memory footprint (4-36KB) to facilitate rapid changes to the selected object ({\raisebox{-0.1ex}{\textasciitilde}\appTime per edit). This approach also facilitates progressive edits. 
For edits involving geometric modifications, we fine-tune the whole NeRF (i.e., both geometry and appearance branches) on the \context, which is iteratively updated to capture the updated geometry (i.e., NeRF density). In the generative editing setup, we modify the image context using Stable Diffusion~\cite{stablediffusion} with conditional controls produced using rendered depth or Canny edge maps via a pre-trained ControlNet~\cite{controlnet}. This allows us to fine-tune on a small set of images (4 images for a 2$\times$2 image-grid context) for each iteration, resulting in accelerated re-training of the NeRF ({\raisebox{-0.1ex}{\textasciitilde}\geoTime for each edit).  
Figure~\ref{fig:overview} shows an overview of our algorithm.

\subsection{\triplanelite: Residual Tri-plane Feature Field}
\label{subsec:triplanelite}
We use a tri-plane \cite{eg3d} to represent a NeRF owing to its efficiency and compactness. We first learn the captured geometry to the tri-plane using an MLP $\phi_{geom}$ to map to density $\sigma$ and geometric features $f_{geom}$. We then map $f_{geom}$ to volumetric semantic features, $f_{sem}$, using an MLP $\phi_{sem}$ with supervision from a large-scale pre-trained model (we have used DINO~\cite{dino} features in our experiments). Based on the intuition that it is possible to learn the color of a point directly from its semantic features, we distill the semantic features using another MLP $\phi_{color}$ to learn color ($r,g,b$). Given a camera, we `render' the final pixel colors and semantic features using volumetric rendering. 
See Figure~\ref{fig:model}. 
As a simplification, we ignore view dependent effects by eliminating the viewing direction as an input to the NeRF, similar to EG3D \cite{eg3d}.

We follow Neural Feature Fusion Fields\cite{tschernezki2022neural} to select the editing region. Once the NeRF is pre-trained on a scene, the user can select an image patch of the object to indicate the region of interest. Based on the 2D query patch mask ($\mathcal{M}^q_{2D}$), we compute the mean of the 2D feature vectors of each pixel of the patch, $\bar{f}_{2D}^{q}$. We calculate similarity distance, $f_{distance} = ||\bar{f}_{2D}^{q} -  f_{sem}||^2$, for each 3D point in volumetric rendering processing. A user can select a threshold value $thr$ in the $f_{distance}$ range. Based on the user selected threshold distance, $thr$, corresponding points are selected as the 3D mask of the object, \mask. For example, we can get the \mask with points whose $f_{distance}$ are lower than the $thr$. Then we can set the density $\sigma$ of unmask/mask points to $0$. See Figure~\ref{fig:selection}.

\begin{figure}[b!]
    \centering
    \includegraphics[width=\columnwidth]{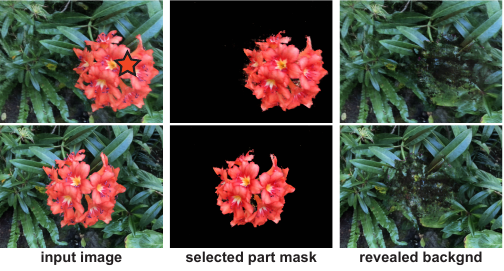}
    \caption{
    Once the input posed images $\{I_i, C_i\}_{i=1:n}$ are feature-distilled into TriplaneLite, the user can select a region in any of the images (shown in orange here), which is then used to extract a 3D mask \mask. Suppressing the corresponding signal in the mask, reveals the background across views. 
    }
    \label{fig:selection}
\end{figure}

\subsection{\Context}
\label{subsec:gridContext}
To edit the NeRF, we need to optimize it on a new set of edited images. While it is possible to make simple color changes while maintaining view consistency, the problem is non-trivial for making large appearance or geometric modifications. This problem is exacerbated for generative models such as Stable Diffusion \cite{stablediffusion}, which can produce markedly different images resulting in geometric inconsistencies in the edited NeRF. To address this challenge, we introduce a simple yet highly effective solution that we term as \context: merging multiple images into a single image grid when editing. After sampling views from the training dataset, we generate a $2\times 2$ grid of images.

This allows the generative model to share the same latent code when modifying the image, resulting in the edited object's coherent appearance and geometry. The solution can be seen as a crude approximation to learning an attention map by giving local image context. This context also applies to traditional image editing software as various operations such as contrast adjustment, can be uniformly applied across multiple images. Furthermore, it simplifies the editing process as the user only needs to make a single modification.

\begin{figure}[t!]
    \centering
\includegraphics[width=\columnwidth]{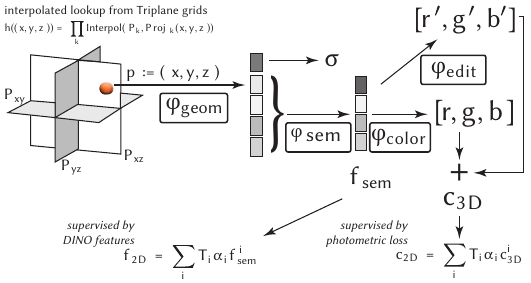}
    \caption{
We encode an object as a NeRF using our 
    TriplaneLite structure that takes as input a point $\mathbf{p}:=(x,y,z)$ and encodes it as features $h(\mathbf{p})$ by projection and interpolation of features from three planar grids ($P_{xy}, P_{yz}, P_{xz}$). We then enable learning via four different MLPs $\phi_{geom}, \phi_{sem}, \phi_{color},$ and $\phi_{edit}$, to factorize density, semantic features, color, and residual appearance respectively. Training is supervised via photometric loss and distillation of image space semantic features (DINO features). 
    This enables semantic selection (see Figure~\ref{fig:selection}). 
    Furthermore, the structuring lets us interactively receive appearance updates while requiring a low memory overhead (36KB/edit).  
    }
    \label{fig:model}
\end{figure}

\subsection{NeRF Editing}
We divide the editing process into two separate workflows depending on the task. 

\subsubsection{Smaller appearance-only edits}
For appearance changes using traditional image processing such as recoloring, contrast changes, altering the white balance, etc., there are many mature image editing tools such as Gimp and Adobe Photoshop. Generative methods such as InstructPix2Pix \cite{brooks2023instructpix2pix} enable more creative appearance changes using natural language as instruction. To edit the NeRF on the updated images, we propose adding a simple 3-layer residual MLP $\phi_{edit}$ with a minimal footprint of 36KB. See Figure~\ref{fig:model}. During re-training on the 2$\times$2 edited \context, we freeze all layers of the NeRF except $\phi_{edit}$ to learn the new color. Essentially, this operation learns a new mapping of the distilled features to the target color values provided by the guidance context images.  Training such a small network, while keeping geometric encoding fixed, converges on merely four images after two epochs, resulting in rapid optimization as, 

\begin{equation}\label{residualmlp}
\begin{split}
c_{2D} &:= \sum_i T_i\alpha_i{c_{3D}}_i \odot (\raisebox{-0.8ex}{\textasciitilde}\mask) + \\
& \sum_i T_i\alpha_i
({c_{3D}}_i + \phi_{edit}({f_{sem}}_i)) \odot \mask.
\end{split}
\end{equation}

\subsubsection{Larger edits}
Larger edits, especially involving geometry changes, can benefit from additional views to denoise the inconsistencies in edited views using generative methods. Therefore, we fine-tune the entire NeRF using an iteratively updated \context. Specifically, we edit the image context using Stable Diffusion and add conditional control to preserve the local properties of the object. We utilize image inpainting to modify the object selectively, i.e., the masked region, along with ControlNet~\cite{controlnet} with depth maps and/or Canny edge maps for guidance. The depth map is trivial to obtain as, after pretraining, the NeRF can accurately represent the scene, and can be obtained by simply volume rendering the sampled density values. See Figure~\ref{fig:depth_outputmetallic_horns }.

Next, we describe how we obtain the iteratively updated \context. Let $I_i^j \in I$ be the original \context and $E_i^j \in E$ be the edited image context, where $i$ is a camera position chosen at random and $j$ is the epoch. For an edited 2$\times$2 image context, $E^0 := \{E_{a}^0, E_{b}^0, E_c^0, E_d^0\}$, after the first epoch, we fix two cells in the grid- $E_{a}^0$ and $E_{b}^0$ without masking to serve as guidance for the subsequent iterations of inpainting for $E_{c}^0$ and $E_{d}^0$ (masked). This simple adaptation allows the generative method to inpaint the unmasked cells (new views) using the previously edited views as references, thereby ensuring coherence in the edited images across epochs. In our experiments, three iterations resulting in 8 edited views prove sufficient.

\begin{figure}[t!]
    \includegraphics[width=\columnwidth]{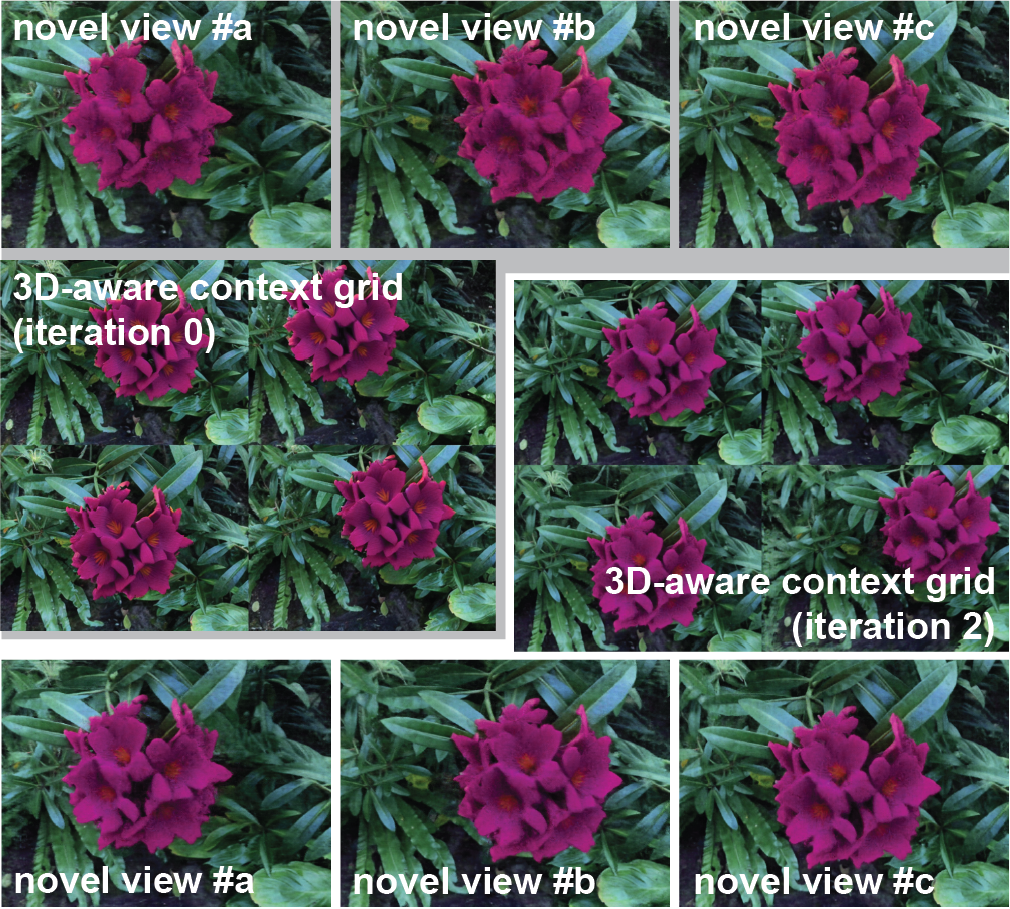}
    \caption{
   Although our \context helps to synchronize edits across nearby views, inconsistencies can still occur. We iterate between context-guided image edits, distillation into a refined NeRF, and regenerating new guidance images. Typically, we found that 2-3 iterations was enough to strike a balance between expressive edits and interactive performance. 
    }
    \label{fig:iterativeUpdate}
\end{figure}

\begin{figure}[t!]
    \includegraphics[width=\columnwidth]{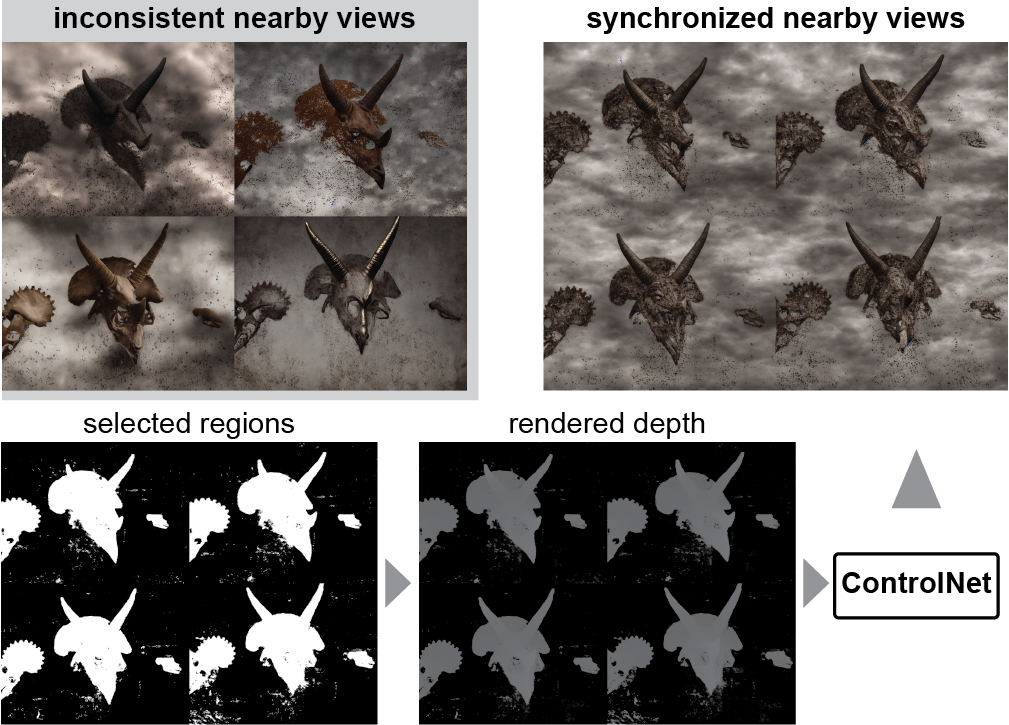}
    \caption{
(Top-left)~Without any 3D guidance, edits from nearby (camera) views can be inconsistent, and hence, any intended edits can get lost during subsequent NeRF refinement. 
Instead, we advocate using \context, a simple change that encourages edit consistency among nearby views. 
Given the selected object mask, 
we extract depth-only rendering on a $2\times 2$ grid of nearby cameras to give guidance to a pre-trained ControlNet~\cite{controlnet}, optionally augmented with feature edges and/or text prompts. (Top-right)~These edits generated using \context are more consistent and preserved during subsequent NeRF fine-tuning. 
}
\label{fig:depth_outputmetallic_horns }
\end{figure}

\subsection{Optimization}
We re-train the NeRF on the edited \context ($E_{i}^j$) using MSE with \textit{L1} regularization \cite{Chen2022ECCV} and {total variation L2} \cite{kplanes_2023} regularization to encourage sparsity and smoothness in the tri-plane. \textit{L1} and {total variation L2} regularization help to reduce the problem of floaters and artifacts during rendering in the edited NeRF, particularly for geometric edits. To further suppress floaters and artifacts, we motivate the rendered depth ($R_{\text{depth}}$) in the masked region to be close to the depth of the edited view ($E_{\text{depth}}$) estimated using DPT Large~\cite{ranftl2021vision}. We also propose an optional mask loss $\mask \odot ||\sigma_{original} - \sigma_{edited}||^2_2$ to penalize changes in density outside the masked region. The total loss, $L$, is the sum of the reconstruction MSE, the \textit{L1} penalty, TV loss ($\mathcal{L}_{T V}$), and the depth loss.

\begin{equation}\label{optimization}
\begin{split}
L := & \frac{1}{mn}\sum^{n}_{q=1} \sum_{p=1}^{m} (R_{p}^q - c_{2Dp}^q)^2 + \lambda_{1} \sum |W_P| +  \lambda_{2} \mathcal{L}_{T V}({W_P})  \\
+ &\lambda_{3}({\mask \odot ||R_{\text{depth}} - E_{\text{depth}}||_2^2} \\
&+ (\raisebox{-0.1ex}{\textasciitilde}\mask) \odot ||R_{\text{depth}}- C_{\text{depth}}||_2^2), 
\end{split}
\end{equation}

\vspace{-1.5pc}
\begin{equation}\label{tv}
\mathcal{L}_{T V}({W_P}) := \sum \left(\left\|W_P^{i, j}-W_P^{i-1, j}\right\|_2^2+\left\|W_P^{i, j}-W_P^{i, j-1}\right\|_2^2\right)
\end{equation}

where $m$ and $n$ are the height and width of the rendered image, respectively; $p \in 1:m$ and $q \in 1:n$ are the pixel locations, $\lambda_1$ and $\lambda_2$ are constants to control the regularization, $\lambda_3$ controls depth loss, $\lambda_4$ controls mask loss, and $W_P$ are the weights of the tri-plane parameters ($P_{xy}$, $P_{xz}$, and $P_{yz}$).

\section{Results}

\begin{figure*}[!t]
  \centering
  \vspace{2pc}
  \includegraphics[width=\textwidth]{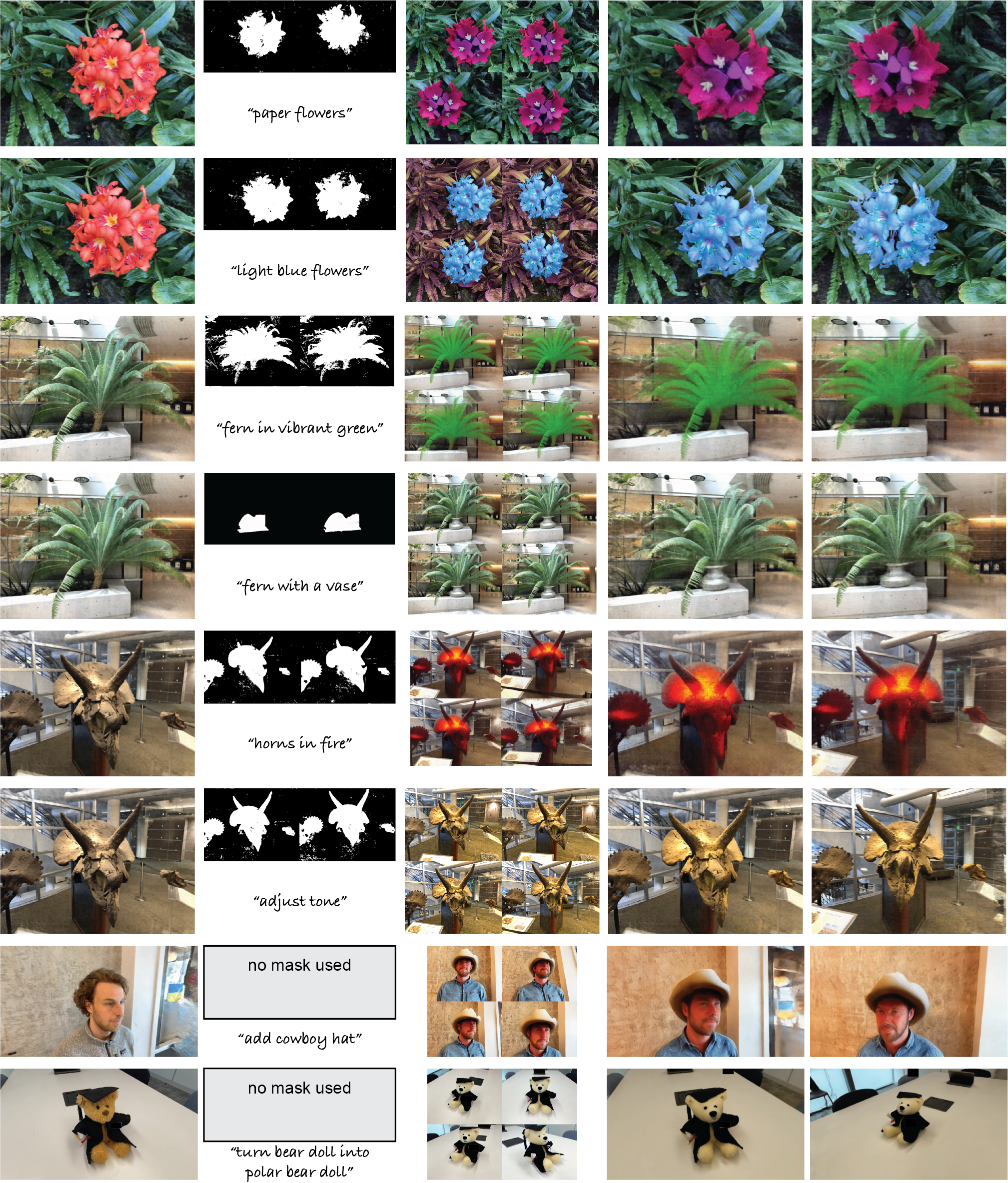}
  \caption{Results gallery.
     (Left-to-right) Input scene; selection masks with target edit description; \context (one iteration shown); final NeRF in source view and a novel view. 
     Please see the supplementary webpage for video results. \\}
  \label{fig:result-1}
\end{figure*}

\begin{table*}[!t]
    \centering
    \caption{Our method is significantly faster than InstructNeRF2NeRF (IN2N) \cite{haque2023instruct} while occupying a fraction of its memory footprint, particularly when utilizing the residual MLP $\phi_{edit}$ for appearance changes. We measure the time required of ours for image size (504 $\times$ 378) and IN2N for image size (497 $\times$ 369). All experiments are run on a single Nvidia A100 GPU.}
    \vspace{-0.5pc}
    \begin{tabular}{ c| c| c| c } 
    Method & Settings & Time $\downarrow$ (minutes)& Storage$\downarrow$ (KB)\\
    \hline \hline

\hfil {IN2N} & 15000 iterations (large edits) & $\sim$ 50 & $\sim$171844 \\ 
\hfil {IN2N w/ 2$\times$ 2 context} & 15000 iterations (large edits) & $\sim$ 18 & $\sim$171844 \\ 
    \hline
    \hfil {Ours Small appearance-only} & 500 iterations (small edits) & $\sim$ 0.25 & 36 \\ 
  
    {\hfil Ours Larger edits} & $\sim$4500 iterations (large edits) & $\sim$ 1.42 & 32917 \\ 
    \end{tabular}
    \label{tab:Quantitative}
\end{table*}

\paragraph*{Dataset.}
To demonstrate the effectiveness of our method, we show experiments on scenes from the LLFF \cite{mildenhall2019llff} and Blender Synthetic \cite{barron2022mip} datasets. We select three scenes from the LLFF dataset -- flower, horns, and fern -- and choose the Lego sequence from the Blender Synthetic dataset. To show the efficacy of our method on casually captured videos, we utilize the face dataset from InstructNeRF2NeRF \cite{haque2023instruct}, and capture a video of a bear doll on iPhone12 Pro using Polycam \cite{polycam} processed with Nerfstudio \cite{nerfstudio}. 

\paragraph*{Implementation details.}
The MLPs- {$\phi_{geom}, \phi_{sem}, \phi_{color}$} have {2, 1, 2} layers respectively with ReLU and sigmoid activation functions. Our $\phi_{edit}$ MLP consists of 3 layers with the Leaky ReLU activation function. We follow K-Planes and TensoRF \cite{kplanes_2023, Chen2022ECCV} for parameter settings; further details will be found in our code During NeRF pretraining, we uniformly sample 96 and 192 rays for the LLFF dataset and Blender Synthetic datasets, respectively. The architecture is optimized using Adam \cite{kingma2014adam} for 40,000 iterations on all views. During editing, for small appearance changes, we train our MLP $\phi_{edit}$ for 500 iterations with a batch size of 1024 for each edited image. For larger edits involving geometric changes, we re-train the NeRF for 1 epoch for edited images with a batch size of 512 and a learning rate of $\num{2e-4}$; As for the regularisation parameters, higher values lead to fewer floaters, but also increase sparsity of density. We provide further parameters details in our code. We test the official implementation (v0.1.0) of InstructNeRF2NeRF using the default settings in Nerfstudio \cite{nerfstudio}. For the bear and face data sets, we used the default settings of Nerfstudio v0.3.2 for pre-training for 30,000 iterations with a batch size of 4096. In addition, we extended the \context to the iterative dataset update method of InstructNeRF2NeRF to update the image every 100 iterations using a 2x2 context for comparison.

\subsection{Layered Editing}
Our lightweight architecture allows a single residual MLP comprising of $\sim$4 - 36KB to store each edit. Thus, we can perform layered editing of NeRFs akin to image editors to enable more controlled and creative workflows. Each MLP essentially learns a operation such as hue change, contrast change, tone mapping, etc. which remaps colors and thus, these MLPs can be chained to produce layered (combined) edits as seen in \Cref{fig:layered}.

\begin{figure}[h]
    \centering
\includegraphics[width=0.65\linewidth]{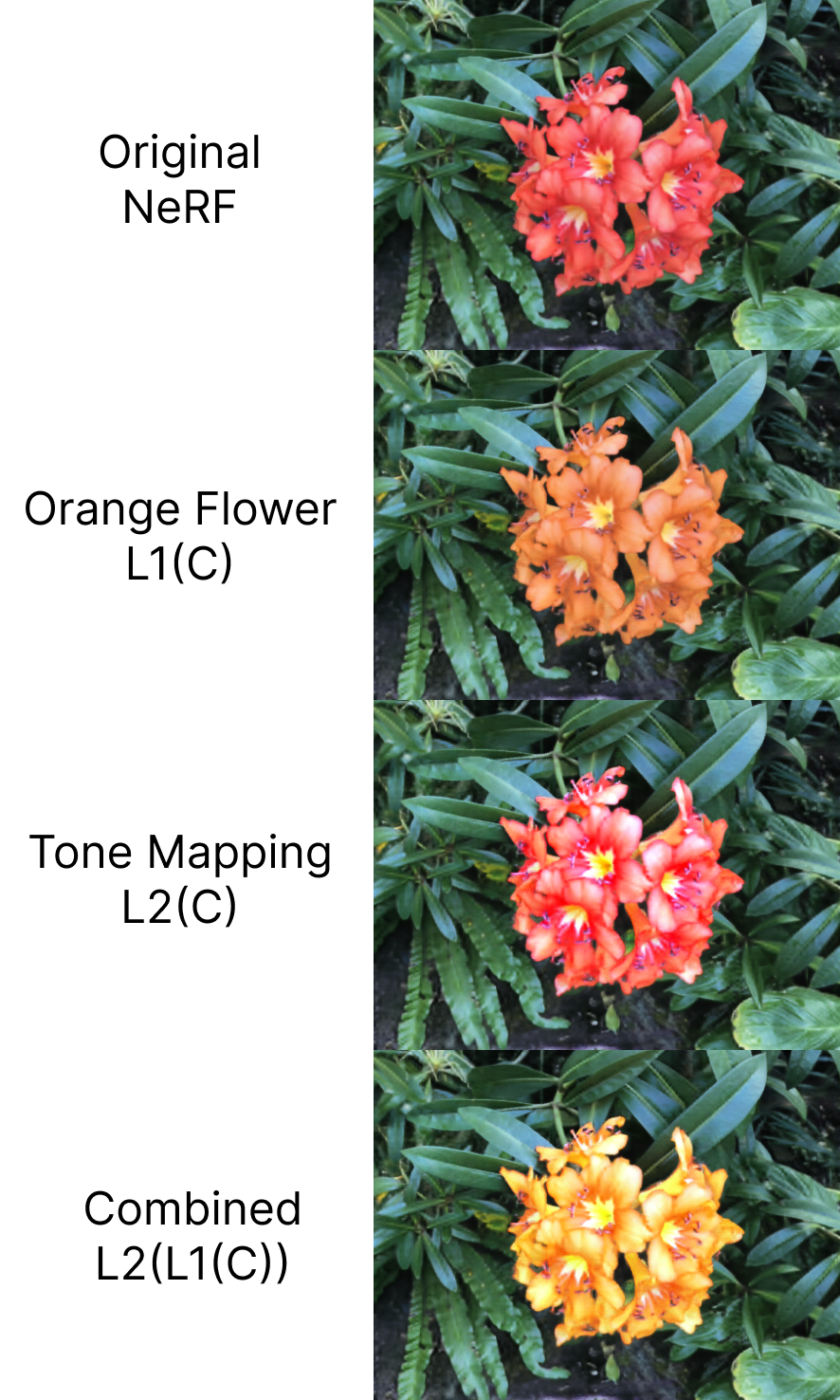}
    \caption{Layered editing. We can chain multiple layers of residual MLP $\phi_{edit_{c2c}}$ to map colors with a minimal memory footprint of 4 KB. Here, $L1(c)$  color change and $L2(c)$ tone mapping are combined to produce the final edit (orange tone mapped flower).}
    \label{fig:layered}
\end{figure}

\subsection{Qualitative Evaluation}
Our method can handle a wide range of edits, including appearance changes, small geometry changes, and addition and deletion of objects, while maintaining a fast and lightweight framework. We compare our method against related methods to recolor a flower scene as illustrated in \Cref{fig:motivation}. We encourage the readers to explore our web page for video comparisons. CLIP-NeRF~\cite{wang2022clip} bleeds the color changes into the global scene causing unwanted recoloring. While the editing is more localized for DFF~\cite{kobayashi2022decomposing}, we see unnatural color gradients. We utilize traditional image editors in our workflow to recolor the \context. Although RecolorNeRF too shows desirable results, its editing framework is less intuitive and limited to simple color changes while being an order of magnitude slower. Our approach demonstrates the efficacy of image processing tools for making precise color changes in NeRFs. This is further highlighted in the accurate contrast tone mapping edits in \Cref{fig:result-1}.

To show larger appearance changes, several examples in \Cref{fig:result-1} show changes in texture, color, and geometry. The creative edits utilizing our generative workflow show substantial appearance changes while being local such as editing the appearance of horns, fern and flower. Our method demonstrates adaptability by accommodating small geometric changes, such as adding a vase in the fern sequence or a cowboy hat in the face scene, or removing the flower (\Cref{fig:selection}). Additionally, we present challenging geometric edits in \Cref{fig:gepmetric-edits}. We show qualitative comparison with InstructNeRF2NeRF in \Cref{fig:design_ablation}.

\begin{figure}[h!]
    \centering
    \includegraphics[width=\linewidth]{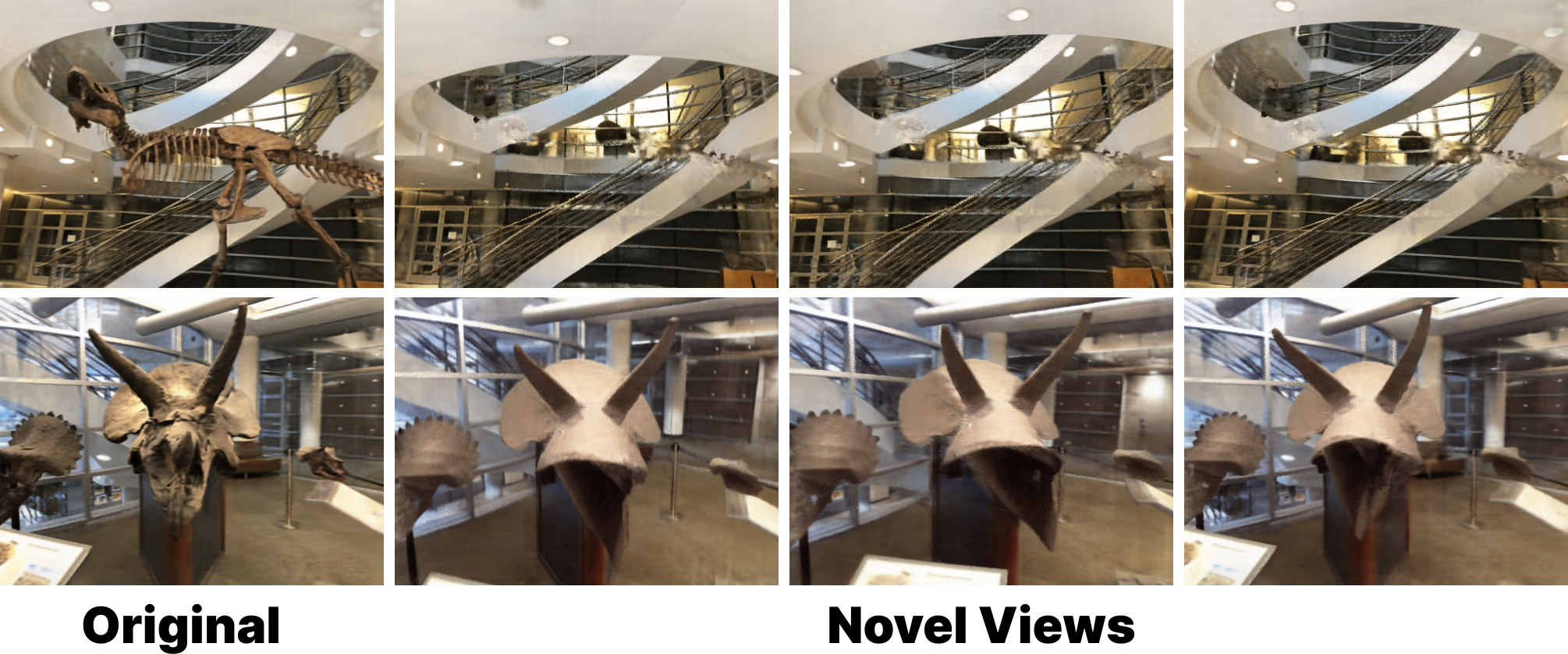}
    \caption{
Geometric edits. Our method is versatile enough to make large deletions in a scene ("remove T-rex") as well as generative appearance edits with accompanying geometric changes ("paper horns").
}
    \label{fig:gepmetric-edits}
\end{figure}

\subsection{Quantitative Evaluation}
Our method is significantly faster than existing works, which can take thirty minutes to an hour for larger appearance changes or geometric modifications. We compare the time and storage required by our method to edit a scene against InstructNeRF2NeRF in \Cref{tab:Quantitative}. For smaller appearance only edits, our method shows $\sim$ 40x speedup over InstructNeRF2NeRF while occupying a fraction of the memory space when training is done on MLP $\phi_{edit}$. The minimal memory footprint of MLP $\phi_{edit}$ (36KB) paves the way for layered editing of NeRFs, which would not only help to make fine-grained changes in the scene but also allow the user to preserve the editing history. For larger edits, our method is $\sim$ 35x faster when using iterative context. 

Editing quality is inherently a subjective opinion. However, we compute CLIP Text-Image Direction Similarity~\cite{brooks2023instructpix2pix} to show that the edited scene is compatible with the text instruction in the CLIP \cite{clip} space. When using our 2$\times$2 \context with InstructNeRF2NeRF~\cite{haque2023instruct} we obtain similar or higher CLIP score than IN2N while requiring significantly lesser editing time and iterations as seen in \Cref{fig:graph1} and \Cref{fig:graph2}.  

\begin{figure}[t!]
  \centering
  \includegraphics[width=\columnwidth]{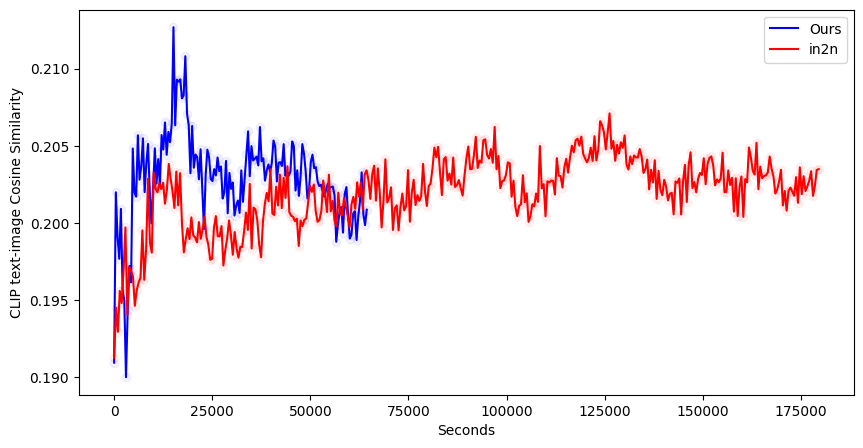}
  \caption{
   Editing time. When using our 2$\times$2 \context with IN2N~\cite{haque2023instruct}, we reduce editing time by a third.
    }
  \label{fig:graph1}
\end{figure}

\begin{figure}[t!]
  \centering
  \includegraphics[width=\columnwidth]{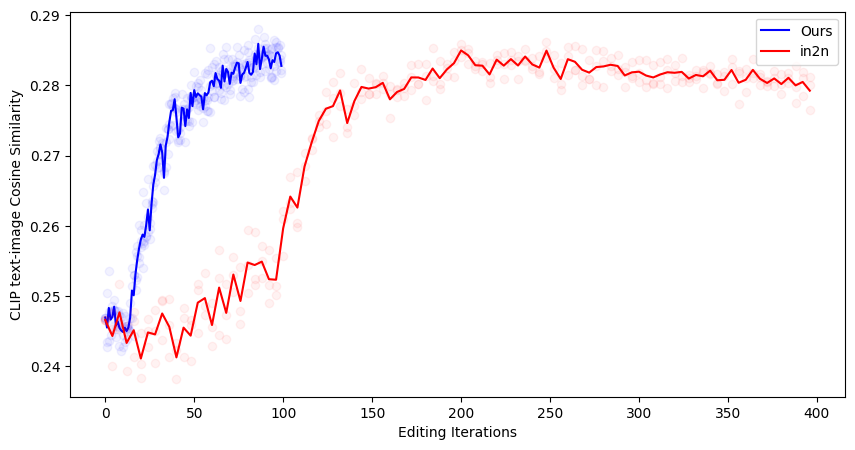}
  \caption{
Editing iterations. When using our 2$\times$2 \context with IN2N~\cite{haque2023instruct}, we reduce the number of editing iterations required by more than a third.
}
    \label{fig:graph2}
\end{figure}

\begin{figure}[t!]
  \centering
  \includegraphics[width=\columnwidth]{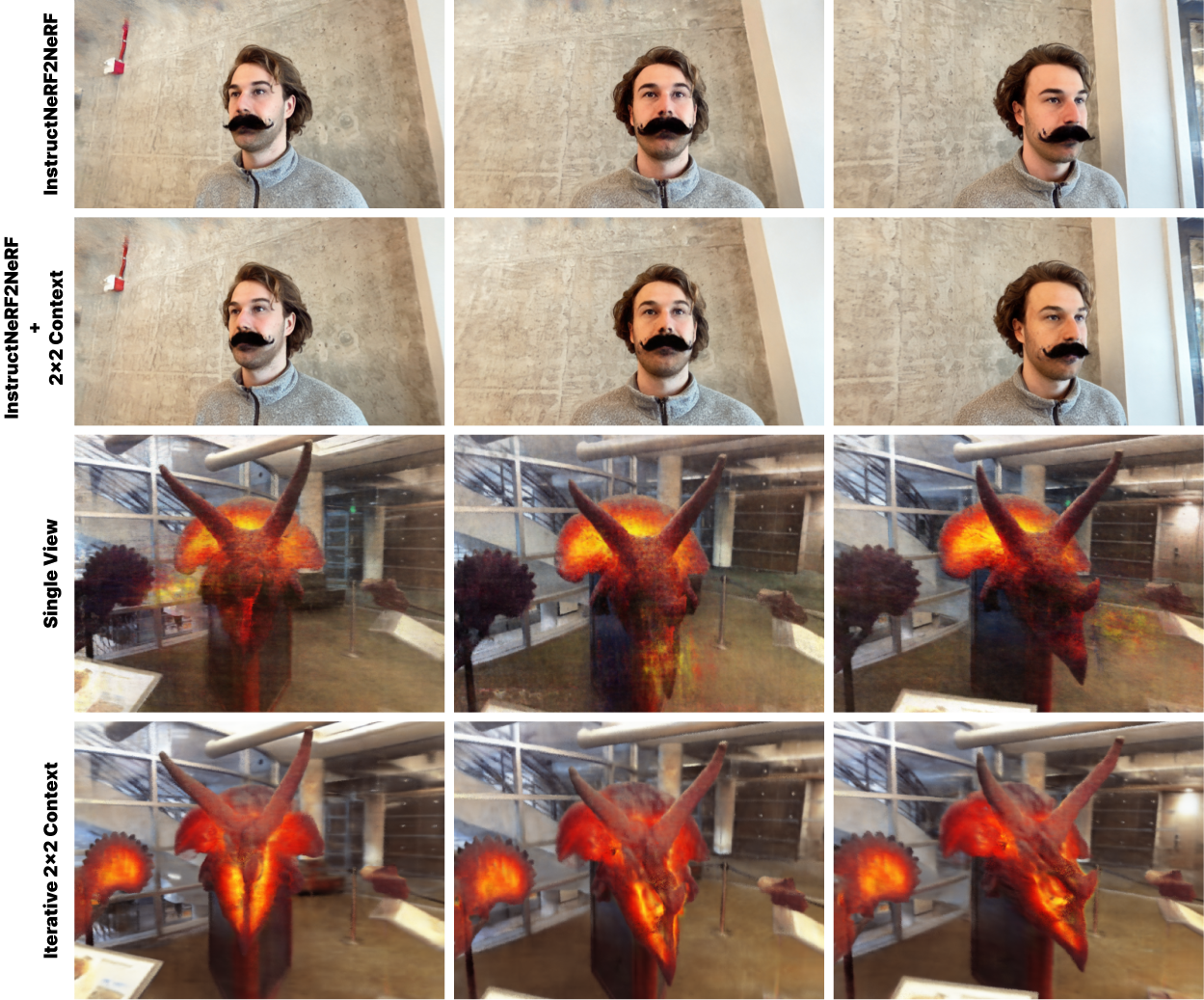}
  \caption{Qualitative comparison. 
    (Top-to-bottom)~InstructNeRF2NeRF~\cite{haque2023instruct}, InstructNeRF2NeRF with 2$\times$2 \context, single view, 2$\times$2 iterative \context. The edited scene using InstructNeRF2NeRF has artifacts around the mustache whereas when using 2$\times$2 \context, it is able to overcome the issue with view consistency. When using single view to retrain the NeRF, the edited scene has a large amount of artifacts whereas when using 2$\times$2 iterative \context, the 3D prior helps to maintain view consistency. Please see the supplementary webpage for additional videos.}
  \label{fig:design_ablation}
\end{figure}

\subsection{Ablation}
We probe the effectiveness of the iterative \context in this section. We compare quantitative and qualitative results of our 2$\times$2 iterative \context against a baseline of editing using a single view (without using any contextual information). When training is done on a single view, performance on novel views is significantly reduced as there is no geometric prior to appropriately map the edited appearance and geometry from a single edited 2D image to 3D. While a fixed \context can ease the problem, iteratively updating the context allows us to provide additional views, enhancing the convergence of the network as seen in \Cref{fig:design_ablation}. This hypothesis is further supported by the results of our user study presented in \Cref{tab:user-study}. The study involved 6 edited scenes and 20 participants, each tasked with choosing between two options (single view vs 2$\times$2 iterative \context)- the video that most accurately matched a provided text prompt (Text-Scene Similarity) and selecting the video that was the most view-consistent and had the least artifacts. An overwhelming 68.33\% of the participants found that edits using \context were more similar to the provided text prompt and 96.67\% of the participants found that using \context led to more view-consistent edits. Utilizing a single view can lead to sufficient alignment of the edited scene with the text prompt and the stochastic nature of edits can affect user preferences, therefore we see a comparatively lower disparity in preference between \context and single-view. However, a lack of 3D awareness in the single view leads to artifacts in the edited scene; hence, we see a larger disparity in user preference for \context.

\begin{table}[h!]
\caption{User study. Majority of the users prefer edits using our 2$\times$2 Iterative \context as it shows higher similarity with the edit instruction and is significantly more view-consistent.}
\begin{tabular}{p{3cm}|p{2cm}|p{2.5cm}}
Method & Text-Scene Similarity & View-consistency (least artifacts \& floaters) \\
\hline \hline
Single view & 31.67\% & 3.33\% \\
2$\times$2 Iterative \context & 68.33\% & 96.67\% \\
\end{tabular} \label{tab:user-study}
\end{table}

\begin{figure}[t!]
    \centering
    \includegraphics[width =\columnwidth]{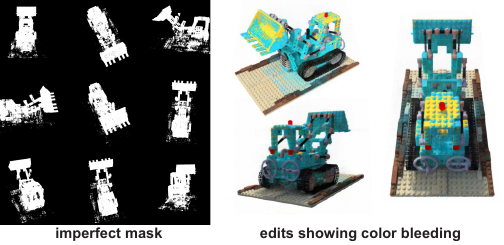}
    \caption{
Our method can produce imperfect mask \mask when DINO based semantic feature distillation fails to capture in-class color variations fully. In addition, specularity contradicts our assumption of a uniform diffuse appearance. Here, the inferior mask results in changes in the Lego scene to produce color bleeding (blue into yellow). One possible solution to address this issue would be to combine feature and color-based selection. 
}
    \label{fig:limitation}
\end{figure}

\section{Conclusions}

We have presented \name as a fast and lightweight framework that supports object-centric NeRF edits at interactive rates. We observe that existing NeRFs can be restructured to facilitate semantic selection via a feature distillation at the training stage, and any subsequent appearance change can then simply be interpreted as a remapping operation from the selected features to the target colors. This greatly simplifies the selection as well as lightweight storage of appearance edits via residual MLPs, enabled by our TriPlaneLite architecture. Further, we introduce editing via a novel \context 
that allows users to leverage existing image editing setups for NeRF editing. 

Our approach has several limitations that we plan to address in future explorations. 

(i)~\textit{Handle larger geometric changes:} While we support only finer geometric changes, we would like to explore how to enable larger geometric changes at interactive rates, without having to re-optimize the NeRF from scratch or distill the NeRF asset into textured meshes.

(ii)~\textit{Support specular effects:} In the future we would like to extend appearance edits to handle view-dependent specular effects as seen in  \Cref{fig:limitation}. One possibility is to create NeRFs with a factorized diffuse and specular representation of color, and then apply appearance edits to only diffuse channels. However, the challenge is to make such a factorization without sacrificing the simplicity of the NeRF capture process. Moreover, it is unclear how to create view-dependent edits using the \context. 

(iii)~\textit{Unifying NeRF editing and generation:} Finally, we would like to link generative image editing and 3D-aware NeRF guidance more closely to explore direct 3D-aware latent code blending (i.e., for latent diffusion), across views, towards a synchronized multi-view/image diffusion model~\cite{stablediffusion,bar2023multidiffusion} for text-guided editing, while maintaining interactive framerates.

\bibliographystyle{alpha-doi} 
\bibliography{reference}

\newcommand{\etalchar}[1]{$^{#1}$}
\begin{thebibliography}{\uppercase{FKMW{\etalchar{*}}23}}

\bibitem[BHE23]{brooks2023instructpix2pix}
\textsc{Brooks T., Holynski A., Efros A.~A.}:
\newblock Instructpix2pix: Learning to follow image editing instructions.
\newblock In \emph{Proceedings of the IEEE/CVF Conference on Computer Vision and Pattern Recognition} (2023), pp.~18392--18402.

\bibitem[BMV{\etalchar{*}}22]{barron2022mip}
\textsc{Barron J.~T., Mildenhall B., Verbin D., Srinivasan P.~P., Hedman P.}:
\newblock Mip-nerf 360: Unbounded anti-aliased neural radiance fields.
\newblock In \emph{Proceedings of the IEEE/CVF Conference on Computer Vision and Pattern Recognition} (2022), pp.~5470--5479.

\bibitem[BTYLD23]{bar2023multidiffusion}
\textsc{Bar-Tal O., Yariv L., Lipman Y., Dekel T.}:
\newblock Multidiffusion: Fusing diffusion paths for controlled image generation.
\newblock \emph{arXiv preprint arXiv:2302.08113} (2023).

\bibitem[CHIS23]{croitoru2023diffusion}
\textsc{Croitoru F.-A., Hondru V., Ionescu R.~T., Shah M.}:
\newblock Diffusion models in vision: A survey.
\newblock \emph{IEEE Transactions on Pattern Analysis and Machine Intelligence} (2023).

\bibitem[CLC{\etalchar{*}}22]{eg3d}
\textsc{Chan E.~R., Lin C.~Z., Chan M.~A., Nagano K., Pan B., De~Mello S., Gallo O., Guibas L.~J., Tremblay J., Khamis S., et~al.}:
\newblock Efficient geometry-aware 3d generative adversarial networks.
\newblock In \emph{Proceedings of the IEEE/CVF Conference on Computer Vision and Pattern Recognition} (2022), pp.~16123--16133.

\bibitem[CTM{\etalchar{*}}21]{dino}
\textsc{Caron M., Touvron H., Misra I., J{\'e}gou H., Mairal J., Bojanowski P., Joulin A.}:
\newblock Emerging properties in self-supervised vision transformers.
\newblock In \emph{Proceedings of the IEEE/CVF international conference on computer vision} (2021), pp.~9650--9660.

\bibitem[CTT{\etalchar{*}}22]{chiang2022stylizing}
\textsc{Chiang P.-Z., Tsai M.-S., Tseng H.-Y., Lai W.-S., Chiu W.-C.}:
\newblock Stylizing 3d scene via implicit representation and hypernetwork.
\newblock In \emph{Proceedings of the IEEE/CVF Winter Conference on Applications of Computer Vision} (2022), pp.~1475--1484.

\bibitem[CXG{\etalchar{*}}22]{Chen2022ECCV}
\textsc{Chen A., Xu Z., Geiger A., Yu J., Su H.}:
\newblock Tensorf: Tensorial radiance fields.
\newblock In \emph{European Conference on Computer Vision (ECCV)} (2022).

\bibitem[FKMW{\etalchar{*}}23]{kplanes_2023}
\textsc{Fridovich-Keil S., Meanti G., Warburg F.~R., Recht B., Kanazawa A.}:
\newblock K-planes: Explicit radiance fields in space, time, and appearance.
\newblock In \emph{CVPR} (2023).

\bibitem[FLNP{\etalchar{*}}24]{fischer2024nerf}
\textsc{Fischer M., Li Z., Nguyen-Phuoc T., Bozic A., Dong Z., Marshall C., Ritschel T.}:
\newblock Nerf analogies: Example-based visual attribute transfer for nerfs.
\newblock \emph{arXiv preprint arXiv:2402.08622} (2024).

\bibitem[FTC{\etalchar{*}}22]{yu_and_fridovichkeil2021plenoxels}
\textsc{{Fridovich-Keil and Yu}, Tancik M., Chen Q., Recht B., Kanazawa A.}:
\newblock Plenoxels: Radiance fields without neural networks.
\newblock In \emph{CVPR} (2022).

\bibitem[GKJ{\etalchar{*}}21]{garbin2021fastnerf}
\textsc{Garbin S.~J., Kowalski M., Johnson M., Shotton J., Valentin J.}:
\newblock Fastnerf: High-fidelity neural rendering at 200fps.
\newblock In \emph{Proceedings of the IEEE/CVF International Conference on Computer Vision} (2021), pp.~14346--14355.

\bibitem[GPAM{\etalchar{*}}20]{goodfellow2020generative}
\textsc{Goodfellow I., Pouget-Abadie J., Mirza M., Xu B., Warde-Farley D., Ozair S., Courville A., Bengio Y.}:
\newblock Generative adversarial networks.
\newblock \emph{Communications of the ACM 63}, 11 (2020), 139--144.

\bibitem[GWHD23]{gong2023recolornerf}
\textsc{Gong B., Wang Y., Han X., Dou Q.}:
\newblock Recolornerf: Layer decomposed radiance field for efficient color editing of 3d scenes.
\newblock \emph{arXiv preprint arXiv:2301.07958} (2023).

\bibitem[HGPR20]{harshvardhan2020comprehensive}
\textsc{Harshvardhan G., Gourisaria M.~K., Pandey M., Rautaray S.~S.}:
\newblock A comprehensive survey and analysis of generative models in machine learning.
\newblock \emph{Computer Science Review 38} (2020), 100285.

\bibitem[HHY{\etalchar{*}}22]{huang2022stylizednerf}
\textsc{Huang Y.-H., He Y., Yuan Y.-J., Lai Y.-K., Gao L.}:
\newblock Stylizednerf: consistent 3d scene stylization as stylized nerf via 2d-3d mutual learning.
\newblock In \emph{Proceedings of the IEEE/CVF Conference on Computer Vision and Pattern Recognition} (2022), pp.~18342--18352.

\bibitem[HJA20]{ho2020denoising}
\textsc{Ho J., Jain A., Abbeel P.}:
\newblock Denoising diffusion probabilistic models.
\newblock \emph{Advances in neural information processing systems 33} (2020), 6840--6851.

\bibitem[HSM{\etalchar{*}}21]{hedman2021baking}
\textsc{Hedman P., Srinivasan P.~P., Mildenhall B., Barron J.~T., Debevec P.}:
\newblock Baking neural radiance fields for real-time view synthesis.
\newblock In \emph{Proceedings of the IEEE/CVF International Conference on Computer Vision} (2021), pp.~5875--5884.

\bibitem[HTE{\etalchar{*}}23]{haque2023instruct}
\textsc{Haque A., Tancik M., Efros A.~A., Holynski A., Kanazawa A.}:
\newblock Instruct-nerf2nerf: Editing 3d scenes with instructions.
\newblock \emph{arXiv preprint arXiv:2303.12789} (2023).

\bibitem[HTS{\etalchar{*}}21]{huang2021learningstyle}
\textsc{Huang H.-P., Tseng H.-Y., Saini S., Singh M., Yang M.-H.}:
\newblock Learning to stylize novel views.
\newblock In \emph{Proceedings of the IEEE/CVF International Conference on Computer Vision} (2021), pp.~13869--13878.

\bibitem[JKK{\etalchar{*}}23]{jambon2023nerfshop}
\textsc{Jambon C., Kerbl B., Kopanas G., Diolatzis S., Drettakis G., Leimk{\"u}hler T.}:
\newblock Nerfshop: Interactive editing of neural radiance fields.
\newblock \emph{Proceedings of the ACM on Computer Graphics and Interactive Techniques 6}, 1 (2023).

\bibitem[JMB{\etalchar{*}}22]{jain2022zero}
\textsc{Jain A., Mildenhall B., Barron J.~T., Abbeel P., Poole B.}:
\newblock Zero-shot text-guided object generation with dream fields.
\newblock In \emph{Proceedings of the IEEE/CVF Conference on Computer Vision and Pattern Recognition} (2022), pp.~867--876.

\bibitem[KB14]{kingma2014adam}
\textsc{Kingma D.~P., Ba J.}:
\newblock Adam: A method for stochastic optimization.
\newblock \emph{arXiv preprint arXiv:1412.6980} (2014).

\bibitem[KLB{\etalchar{*}}23]{kuang2023palettenerf}
\textsc{Kuang Z., Luan F., Bi S., Shu Z., Wetzstein G., Sunkavalli K.}:
\newblock Palettenerf: Palette-based appearance editing of neural radiance fields.
\newblock In \emph{Proceedings of the IEEE/CVF Conference on Computer Vision and Pattern Recognition} (2023), pp.~20691--20700.

\bibitem[KMS22]{kobayashi2022decomposing}
\textsc{Kobayashi S., Matsumoto E., Sitzmann V.}:
\newblock Decomposing nerf for editing via feature field distillation.
\newblock \emph{Advances in Neural Information Processing Systems 35} (2022), 23311--23330.

\bibitem[KRWM22]{ReluField_sigg_22}
\textsc{Karnewar A., Ritschel T., Wang O., Mitra N.}:
\newblock Relu fields: The little non-linearity that could.
\newblock In \emph{ACM SIGGRAPH 2022 Conference Proceedings} (New York, NY, USA, 2022), SIGGRAPH '22, Association for Computing Machinery.
\newblock URL: \url{https://doi.org/10.1145/3528233.3530707}, \href {https://doi.org/10.1145/3528233.3530707} {\path{doi:10.1145/3528233.3530707}}.

\bibitem[KW13]{kingma2013auto}
\textsc{Kingma D.~P., Welling M.}:
\newblock Auto-encoding variational bayes.
\newblock \emph{arXiv preprint arXiv:1312.6114} (2013).

\bibitem[LK23]{lee2023ice}
\textsc{Lee J.-H., Kim D.-S.}:
\newblock Ice-nerf: Interactive color editing of nerfs via decomposition-aware weight optimization.
\newblock In \emph{Proceedings of the IEEE/CVF International Conference on Computer Vision} (2023), pp.~3491--3501.

\bibitem[MESK22]{muller2022instant}
\textsc{M{\"u}ller T., Evans A., Schied C., Keller A.}:
\newblock Instant neural graphics primitives with a multiresolution hash encoding.
\newblock \emph{ACM Transactions on Graphics (ToG) 41}, 4 (2022), 1--15.

\bibitem[MRP{\etalchar{*}}23]{metzer2023latent}
\textsc{Metzer G., Richardson E., Patashnik O., Giryes R., Cohen-Or D.}:
\newblock Latent-nerf for shape-guided generation of 3d shapes and textures.
\newblock In \emph{Proceedings of the IEEE/CVF Conference on Computer Vision and Pattern Recognition} (2023), pp.~12663--12673.

\bibitem[MSOC{\etalchar{*}}19]{mildenhall2019llff}
\textsc{Mildenhall B., Srinivasan P.~P., Ortiz-Cayon R., Kalantari N.~K., Ramamoorthi R., Ng R., Kar A.}:
\newblock Local light field fusion: Practical view synthesis with prescriptive sampling guidelines.
\newblock \emph{ACM Transactions on Graphics (TOG)} (2019).

\bibitem[MST{\etalchar{*}}20]{mildenhall2020nerf}
\textsc{Mildenhall B., Srinivasan P.~P., Tancik M., Barron J.~T., Ramamoorthi R., Ng R.}:
\newblock Nerf: Representing scenes as neural radiance fields for view synthesis.
\newblock In \emph{ECCV} (2020).

\bibitem[NPLX22]{nguyen2022snerf}
\textsc{Nguyen-Phuoc T., Liu F., Xiao L.}:
\newblock Snerf: stylized neural implicit representations for 3d scenes.
\newblock \emph{arXiv preprint arXiv:2207.02363} (2022).

\bibitem[PD84]{porter1984compositing}
\textsc{Porter T., Duff T.}:
\newblock Compositing digital images.
\newblock In \emph{Proceedings of the 11th annual conference on Computer graphics and interactive techniques} (1984), pp.~253--259.

\bibitem[PJBM22]{poole2022dreamfusion}
\textsc{Poole B., Jain A., Barron J.~T., Mildenhall B.}:
\newblock Dreamfusion: Text-to-3d using 2d diffusion.
\newblock \emph{arXiv} (2022).

\bibitem[Pol23]{polycam}
\textsc{Polycam}:
\newblock Polycam- lidar and 3d scanner for iphone and android., 2023.

\bibitem[PTL{\etalchar{*}}23]{pan2023drag}
\textsc{Pan X., Tewari A., Leimk{\"u}hler T., Liu L., Meka A., Theobalt C.}:
\newblock Drag your gan: Interactive point-based manipulation on the generative image manifold.
\newblock In \emph{ACM SIGGRAPH 2023 Conference Proceedings} (2023), pp.~1--11.

\bibitem[RAR{\etalchar{*}}22]{nvos}
\textsc{Ren Z., Agarwala A., Russell B., Schwing A.~G., Wang O.}:
\newblock Neural volumetric object selection.
\newblock In \emph{Proceedings of the IEEE/CVF Conference on Computer Vision and Pattern Recognition} (2022), pp.~6133--6142.

\bibitem[RBK21]{ranftl2021vision}
\textsc{Ranftl R., Bochkovskiy A., Koltun V.}:
\newblock Vision transformers for dense prediction.
\newblock In \emph{Proceedings of the IEEE/CVF international conference on computer vision} (2021), pp.~12179--12188.

\bibitem[RBL{\etalchar{*}}22]{stablediffusion}
\textsc{Rombach R., Blattmann A., Lorenz D., Esser P., Ommer B.}:
\newblock High-resolution image synthesis with latent diffusion models.
\newblock In \emph{Proceedings of the IEEE/CVF conference on computer vision and pattern recognition} (2022), pp.~10684--10695.

\bibitem[RDN{\etalchar{*}}22]{ramesh2022hierarchical}
\textsc{Ramesh A., Dhariwal P., Nichol A., Chu C., Chen M.}:
\newblock Hierarchical text-conditional image generation with clip latents.
\newblock \emph{arXiv preprint arXiv:2204.06125 1}, 2 (2022), 3.

\bibitem[RKH{\etalchar{*}}21a]{radford2021learning}
\textsc{Radford A., Kim J.~W., Hallacy C., Ramesh A., Goh G., Agarwal S., Sastry G., Askell A., Mishkin P., Clark J., et~al.}:
\newblock Learning transferable visual models from natural language supervision.
\newblock In \emph{International conference on machine learning} (2021), PMLR, pp.~8748--8763.

\bibitem[RKH{\etalchar{*}}21b]{clip}
\textsc{Radford A., Kim J.~W., Hallacy C., Ramesh A., Goh G., Agarwal S., Sastry G., Askell A., Mishkin P., Clark J., et~al.}:
\newblock Learning transferable visual models from natural language supervision.
\newblock In \emph{International conference on machine learning} (2021), PMLR, pp.~8748--8763.

\bibitem[SCD{\etalchar{*}}23]{song2023blending}
\textsc{Song H., Choi S., Do H., Lee C., Kim T.}:
\newblock Blending-nerf: Text-driven localized editing in neural radiance fields.
\newblock In \emph{Proceedings of the IEEE/CVF International Conference on Computer Vision (ICCV)} (October 2023), pp.~14383--14393.

\bibitem[SSC22]{SunSC22}
\textsc{Sun C., Sun M., Chen H.}:
\newblock Direct voxel grid optimization: Super-fast convergence for radiance fields reconstruction.
\newblock In \emph{CVPR} (2022).

\bibitem[TLLV22a]{tschernezki2022neural}
\textsc{Tschernezki V., Laina I., Larlus D., Vedaldi A.}:
\newblock Neural feature fusion fields: 3d distillation of self-supervised 2d image representations.
\newblock In \emph{2022 International Conference on 3D Vision (3DV)} (2022), IEEE, pp.~443--453.

\bibitem[TLLV22b]{tschernezki22neural}
\textsc{Tschernezki V., Laina I., Larlus D., Vedaldi A.}:
\newblock {Neural Feature Fusion Fields}: {3D} distillation of self-supervised {2D} image representations.
\newblock In \emph{Proceedings of the International Conference on {3D} Vision (3DV)} (2022).

\bibitem[TMS{\etalchar{*}}22]{2022eccvtutorialEncoding}
\textsc{Tancik M., Mildenhall B., Srinivasan P., Barron J., Kanazawa A.}:
\newblock Nerf tutorial eccv 2022, 2022.
\newblock URL: \url{https://sites.google.com/berkeley.edu/nerf-tutorial/home}.

\bibitem[TMT{\etalchar{*}}23]{tsalicoglou2023textmesh}
\textsc{Tsalicoglou C., Manhardt F., Tonioni A., Niemeyer M., Tombari F.}:
\newblock Textmesh: Generation of realistic 3d meshes from text prompts.
\newblock \emph{arXiv preprint arXiv:2304.12439} (2023).

\bibitem[TWN{\etalchar{*}}23]{nerfstudio}
\textsc{Tancik M., Weber E., Ng E., Li R., Yi B., Kerr J., Wang T., Kristoffersen A., Austin J., Salahi K., Ahuja A., McAllister D., Kanazawa A.}:
\newblock Nerfstudio: A modular framework for neural radiance field development.
\newblock In \emph{ACM SIGGRAPH 2023 Conference Proceedings} (2023), SIGGRAPH '23.

\bibitem[WCH{\etalchar{*}}22]{wang2022clip}
\textsc{Wang C., Chai M., He M., Chen D., Liao J.}:
\newblock Clip-nerf: Text-and-image driven manipulation of neural radiance fields.
\newblock In \emph{Proceedings of the IEEE/CVF Conference on Computer Vision and Pattern Recognition} (2022), pp.~3835--3844.

\bibitem[WJC{\etalchar{*}}22]{wang2022nerf}
\textsc{Wang C., Jiang R., Chai M., He M., Chen D., Liao J.}:
\newblock Nerf-art: Text-driven neural radiance fields stylization.
\newblock \emph{arXiv preprint arXiv:2212.08070} (2022).

\bibitem[WSW21]{wang2021generative}
\textsc{Wang Z., She Q., Ward T.~E.}:
\newblock Generative adversarial networks in computer vision: A survey and taxonomy.
\newblock \emph{ACM Computing Surveys (CSUR) 54}, 2 (2021), 1--38.

\bibitem[WZAS23]{wang2023inpaintnerf360}
\textsc{Wang D., Zhang T., Abboud A., S{\"u}sstrunk S.}:
\newblock Inpaintnerf360: Text-guided 3d inpainting on unbounded neural radiance fields.
\newblock \emph{arXiv preprint arXiv:2305.15094} (2023).

\bibitem[XH22]{xu2022deforming}
\textsc{Xu T., Harada T.}:
\newblock Deforming radiance fields with cages.
\newblock In \emph{European Conference on Computer Vision} (2022), Springer, pp.~159--175.

\bibitem[YLT{\etalchar{*}}21]{yu2021plenoctrees}
\textsc{Yu A., Li R., Tancik M., Li H., Ng R., Kanazawa A.}:
\newblock Plenoctrees for real-time rendering of neural radiance fields.
\newblock In \emph{Proceedings of the IEEE/CVF International Conference on Computer Vision} (2021), pp.~5752--5761.

\bibitem[ZA23]{controlnet}
\textsc{Zhang L., Agrawala M.}:
\newblock Adding conditional control to text-to-image diffusion models.
\newblock \emph{arXiv preprint arXiv:2302.05543} (2023).

\bibitem[ZWL{\etalchar{*}}23]{zhuang2023dreameditor}
\textsc{Zhuang J., Wang C., Liu L., Lin L., Li G.}:
\newblock Dreameditor: Text-driven 3d scene editing with neural fields.
\newblock \emph{arXiv preprint arXiv:2306.13455} (2023).

\end{thebibliography}

\end{document}